\documentclass[a4paper,conference]{IEEEtran}
\usepackage{cite}

\usepackage{xcolor}
\usepackage{url}

\ifCLASSINFOpdf
   \usepackage[pdftex]{graphicx}
\else
\fi
\usepackage{tikz}
\usetikzlibrary{arrows,decorations.markings}
\usetikzlibrary{shapes.geometric}
\usetikzlibrary{positioning}
\usetikzlibrary{shapes}

\begin{document}
%
\title{Understanding attention-based encoder-decoder networks: a case study with chess scoresheet recognition}

\author{\IEEEauthorblockN{Sergio Y. Hayashi and Nina S. T. Hirata}
\IEEEauthorblockA{Institute of Mathematics and Statistics -- University of S\~{a}o Paulo\\
S\~{a}o Paulo, Brazil}
}

\maketitle

\begin{abstract}
Deep neural networks are largely used for complex prediction tasks. There is plenty of empirical evidence of their successful end-to-end training for a diversity of tasks. Success is often measured based solely on the final performance of the trained network, and explanations on when, why and how they work are less emphasized. In this paper we study encoder-decoder recurrent neural networks with attention mechanisms for the task of reading handwritten chess scoresheets. Rather than prediction performance, our concern is to better understand how learning occurs in these type of networks. We characterize the task in terms of three subtasks, namely input-output alignment, sequential pattern recognition, and handwriting recognition, and experimentally investigate which factors affect their learning. We identify competition, collaboration and dependence relations between the subtasks, and argue that such knowledge might help one to better balance factors to properly train a network.
\end{abstract}


%
\IEEEpeerreviewmaketitle

\section{Introduction}

Deep learning techniques enabled a significant advance of the state of the art regarding machine learning capabilities. In many fields, previously employed manually designed complex pipelines have been replaced with end-to-end trainable neural networks~\cite{Goodfellow-et-al-2016,zhang2021dive}. There is much evidence showing that with sufficiently large amount of training data and proper adjustments in the network architecture and training strategies, it is possible to make neural networks perform well on a wide range of tasks. However, a lot of the burden has been shifted to data preparation and model choice, setup and training. Without informed choice, these can quickly turn into time and resource consuming tasks.

We started this work interested in automated chess scoresheet reading, where the goal is to process the scoresheet image and generate the sequence of moves written in it. This, together with the desire to explore deep learning based methods, naturally led us to image-to-sequence problems. 
As we found no end-to-end trainable solutions for our particular problem, based on recent solutions proposed for image-to-sequence problems~\cite{xu2015show,you2016image,bluche2017scan,kang2018convolve}, we chose to work with encoder-decoder recurrent networks with attention mechanisms.


Our initial training dataset, built from scoresheets collected at a chess tournament, consisted of 378 instances. With the chosen model and the rather small dataset of ours, we started some exploratory training, which resulted in strong overfitting as expected. We then employed data augmentation strategies to increase the amount of training data, adding images of scoresheets filled with transcriptions of archived game sequences and artificial images generated using script-like fonts that mimic handwriting. After some attempts, training without overfitting was achieved when using a total of five thousand training instances. Details are presented in Section~\ref{sec:preliminaries}.


Following from this point, one could invest efforts to increase the training set until a high recognition rate is achieved. Instead, as the exploratory process revealed some interesting trade-off between some factors, we chose to address the two following issues.

First, besides the increased amount of training data, we had no clues regarding what exactly made training convergence without overfitting to occur, or more precisely, what prevented the network from learning properly in our attempts with fewer training data. Thus we turned our attention to better understand the inner working of these type of architecture. In particular, we look for explanations beyond the just simple and well known ``insufficient amount of training data'' type of answer. We find out that decomposing the task into three underlying subtasks, namely the correct alignment of input and output, the recognition of sequential patterns, and recognition of handwriting itself, provides the means to explain which factors affect learning. We experimentally investigate how learning of these subtasks are affected by some factors and how learning of a subtask affects learning of other subtasks. This analysis shows that properly balancing the factors is key for effective training (Section~\ref{sec:main}).

Second, despite having achieved training convergence without overfitting, performance on test data was low. Thus, we also explore how accuracy can be further increased based on an incremental training approach, without additional effort to enrich the training set. By training incrementally, we managed to boost accuracy on sequences of length 16 from 51.53\% to 79.27\% (Section~\ref{sec:incremental}).


The reported results and discussions are intended to contribute to a better understanding of networks of the type studied in this work. Conclusions and future work are presented in Section~\ref{sec:conclusions}.



\section{Preliminaries}
\label{sec:preliminaries}

\subsection{Related work}
The problem of reading handwritten chess scoresheets can be naturally associated to handwritten text recognition (HTR) problems. A scoresheet reading system should be able to identify the reading start point, reading direction (from left to right and then top to bottom) as well as the stopping point, detect the delimitation of the words and recognize them, to produce a sequence of words (or a sequence of characters). In this sense, these subtasks are similar to the ones for paragraph (multiple lines of text) reading. Scoresheet reading is simpler than general HTR tasks since the number of words in each line is fixed (a move of the first player followed by a move of the second player). In fact, in~\cite{Eicher:2021} a traditional multi-step processing pipeline, where the task is decomposed into multiple substasks (detection of the tabular structure of the form, identification and ordering of the individual cells, segmentation and recognition of the words within each cell, and finally assembling the output sequence with the recognized words) is proposed. We found no end-to-end approaches for this problem.

In the literature, one of the first works using recurrent neural network (RNN) for HTR was proposed in~\cite{graves2006connectionist}. It introduced a layer called connectionist temporal classification (CTC) to align the sequence generated by the RNN to the target sequence in an optimal way, eliminating the need of segmenting characters beforehand. Its application was further investigated using distinct RNN models~\cite{gravesPAMI,bluche2015,wigington2017}. However, CTC limits the applicability to inputs with linear sequences only, such as words or one line of text.

A second wave of methods for HTR used attention mechanisms~\cite{bluche2017scan,kang2018convolve}. The model proposed in~\cite{bluche2017scan} is able to recognize paragraphs but it requires the encoding part of the model to be pre-trained with CTC loss. On the other hand, the model in~\cite{kang2018convolve} does not use CTC, language model nor lexicon, components often used in HTR problems, but its application was demonstrated only for word recognition. Attention~\cite{vaswani2017attention} has been employed also in other tasks such as image captioning~\cite{xu2015show,you2016image} and recognition of handwritten mathematical formulas~\cite{zhang2017watch}. 

More recently, in~\cite{kang2020pay} a model without recurrence, inspired by transformers, has been proposed. In~\cite{coquenet2021span}, the authors propose a model for reading text at paragraph level, learned end-to-end. They manage to implement an implicit line segmentation within the network and then use CTC over the concatenated lines.

For the problem considered here, recent networks are overly complex. Rather, as the capability of encoder-decoder RNNs with attention mechanism in recognizing text at word, line or paragraph levels, without prior image segmentation, has been already demonstrated~\cite{poulos2017characterbased,kang2018convolve,bluche2017scan}, it seems reasonable to assume it is a suitable model to be explored in our case.

\subsection{Dataset}
\label{sec:dataset}

In chess games, players fill chess scoresheets writing the sequence of moves in Standard Algebraic Notation\footnote{SAN: see {\url{https://en.wikipedia.org/wiki/Algebraic_notation_(chess)}}} (SAN). The sequence of moves written in the scoresheets is then recorded in text format, in PGN (Portable Game Notation) format. A PGN file contains the sequence of moves of a game in algebraic notation and additionally some meta data (such as the name of the players, date, final result). An example of the first 20 moves recorded in a PGN file is presented below:

\begin{center}
\begin{minipage}{0.98\linewidth}
  \smallskip
  \centering
  \scalebox{0.8}{\texttt{1. e4 e5 2. Nf3 Nc6 3. Bb5 a6 4. Ba4 Nf6 5. O-O Be7}}
  
  \scalebox{0.8}{\texttt{6. Ke1 b5 7. Bb3 d6 8. c3 O-O 9. h3 Nb8 10. d4 Nbd7}}

  \smallskip
\end{minipage}
\end{center}

For example, the notation \texttt{Nc6} indicates the movement of a knight (\texttt{N}) to position \texttt{c6} (column \texttt{c} and row \texttt{6}) on the board. Notation may differ slightly depending on the language. For instance, \texttt{N} (knight) is denoted as \texttt{C} (``Cavalo'') in Portuguese. Nevertheless, in PGN the moves are always recorded in English notation. Thus, the move \texttt{Cc3} in Portuguese would be recorded as \texttt{Nc3} in the PGN file. Given an image of a filled scoresheet and its respective PGN file, we can straightforwardly create a training instance having the image as the input and the sequence of moves, extracted from the PGN file (without the row numbers), as the target output.

Our dataset consists of 492 images scanned from scoresheets collected at a tournament
in Brazil plus scoresheets with transcription of real game sequences written by volunteers, complemented with artificially generated images. 
Scoresheets with transcription contains game sequences extracted from {\small \texttt{Kaggle.com - 35 Million Chess Games}}\footnote{\url{https://www.kaggle.com/milesh1/35-million-chess-games/version/1}}, converted to Portuguese notation. The artificial images were generated by making collages of individual move images which were either extracted from the scoresheet images or synthesized using handwriting-like fonts. We used a fixed set of individual move images and whenever a move needed to be included in a collage, an instance of that move was chosen randomly from the set. This may result in repeated move images, but on the other hand it allows the creation of as many as needed game sequence image instances.

\subsection{Recognition task characterization}

\paragraph*{Input} Rather than working with the full  image, we decided to restrict the input image to the region corresponding to the first $N$ rows (thus to the first $2N$ moves). This was motivated by the desire to keep experimentation time at acceptable limits and by our judgment that the important layout structures to be recognized are already well represented in just few lines of the scoresheet. 

\paragraph*{Target} Our target is the sequence of moves written in the scoresheet. Recognition of a written move can be modeled as a problem of recognizing each of the individual characters that form the word, or as a word recognition problem. We chose the latter and thus the reading problem is modeled as a prediction of a sequence of words. This seemed to be a reasonable choice since the set of possible moves (vocabulary) is rather small compared, for instance, with any natural language. In addition, this modelling assures that all predicted words will certainly be a valid word (closed lexicon). Furthermore, by considering words as units to be predicted, we allow the model to abstract patterns of sequential moves (language model). In our implementation, the ground-truth is a sequence of codes, each corresponding to a move in the sequence, and thus the language in which the moves are written is not relevant as long as they are correctly mapped to the corresponding code.

\paragraph*{Evaluation} Overall, we evaluate (1) training and validation loss and accuracy, (2) test accuracy, and (3) visual inspection of the attention map. These are shown ahead, for instance in Figure~\ref{fig:experiments}.

\subsection{Model and training setup}
\label{sec:setup}

This section describes the network and training configuration that led to a successful training. By successful training we mean one where we observe a smooth decrease of the loss function until a convergence criteria is met, and one that does not present overfitting (meaning that the validation loss is very similar to the training loss).

\begin{figure}[htb]
    \centering
    \includegraphics[width=0.4\textwidth]{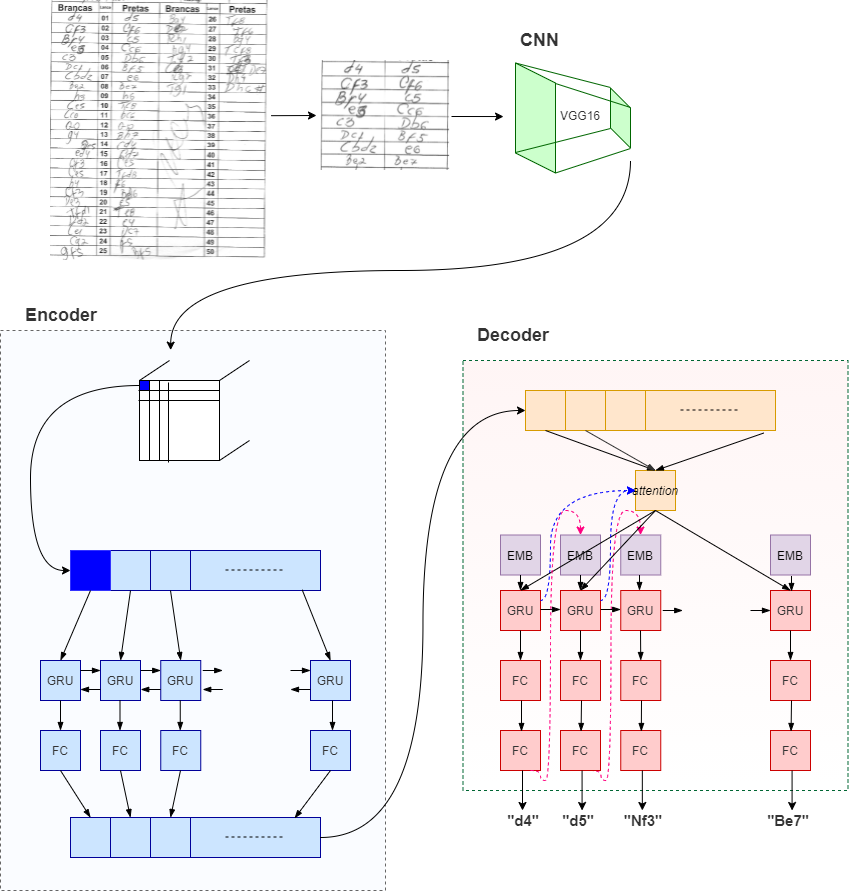}
    \caption{Model architecture overview.}
    \label{fig_model}
\end{figure}

\paragraph*{Architecture}
The architecture of the network is shown in Figure~\ref{fig_model}. A cropped region of an input image goes through the convolutional layers of a pre-trained VGG16~\cite{simonyan2014very} and the final feature map is flattened and sent to a recurrent network consisting of an encoder-decoder structure with attention mechanism. The architecture was based on the implementation of the one proposed in~\cite{xu2015show}, available at the tutorial pages of TensorFlow\footnote{\url{https://www.tensorflow.org/tutorials/text/image_captioning}}. An important modification\footnote{Code is available at \url{https://github.com/sergiohayashi/chess-attention.thesisHayashi2021}} is the inclusion of an encoder RNN in order to capture positional information regarding the sequence in the input image (in image captioning, relational information is important but there is no need for sequential positioning). We use GRU nodes~\cite{cho2014properties} in the RNNs. The internal representation of the nodes in the recurrent layers, as well as the internal fully connected layers of the attention layer, has a dimension of 512. The embedding layer in the decoder has a dimension of 256.

\paragraph*{Training and test data}
Out of the 492 tournament scoresheets, 114 were separated as test set. For training/validation of the model, we used 5000 images (the remainder of tournament scoresheets, 2010 of transcribed ones, and artificially generated ones). For training, we use a 0.8/0.2 partition for training and validation. We note that, input images are regions comprising the first $N=8$ rows of the scoresheet, thus corresponding to the first 16 moves of a game, with dimension in pixels equal to 800x862. This roughly corresponds to 6 attention points per line of the scoresheet.
The vocabulary size regarding the first 16 moves was computed based on a preliminary analysis done on a subset of the images. We use a vocabulary consisting of 175 moves plus 4 special words. When a move is not in the vocabulary, we just map it to the special word denoted \texttt{UNK}.

\paragraph*{Training parameters}
For optimization, we used categorical cross-entropy with logits as the loss function and Adam optimizer, with learning rate of 0.0005. We also use dropout layers with a rate of 0.2 both in the encoder and the decoder RNNs. The batch size was fixed to 16. In order to speed up experimentation time, we set up a somewhat loose convergence criterion. We considered that a training converged when a loss of 0.25 or an accuracy of 0.9 on training data was achieved. We also employed a strategy known as teacher forcing~\cite{lamb2016professor} for training RNNs: during training, rather than using the output predicted by the decoder in the previous time step, the ground-truth output is used, so that wrong prediction of the previous instance is not propagated. 

\section{Model understanding}
\label{sec:main}

The architecture and training setup described above was achieved through some exploratory experimentation. During the exploratory step, we observed that different setup affected training in different ways. In this section we first describe some experiments that demonstrate contrasting results as we change some factors in the training setup, and then next we discuss them.

The discussion we present takes into consideration a decomposition of the task into three subtasks: (1) {\bf Alignment} -- this relates to where the focus of attention is on the input image at a given step; when reading the $i$-th word in the sequence the focus of attention should be precisely on the region where the writing of the $i$-th move is in the image; (2) {\bf Predictability} -- as in any board game, the next move depends on what happened in the game up to that moment. In other words, given the sequence of previous moves, it is possible to associate a probability to the next possible moves, which leads to predictability. This has strong relation with context, and it is often called language model in HTR; (3) {\bf Recognition} -- this relates to the handwriting recognition properly said, based on visual information and eventually on context information.
\begin{figure}
  \begin{minipage}{0.221\textwidth}
     \centering 
     \includegraphics[width=\linewidth]{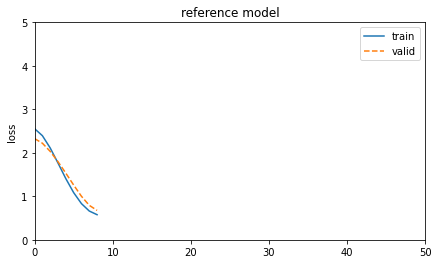}\\
     \includegraphics[width=\linewidth]{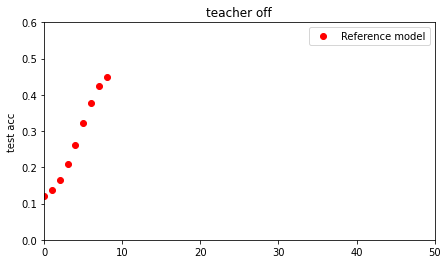}\\
     \includegraphics[width=0.6\linewidth]{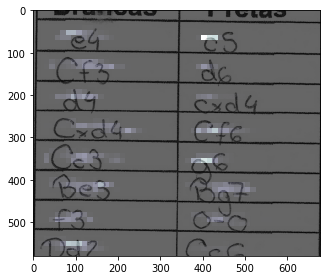}

     {\scriptsize (a) Reference setup}
  \end{minipage} \hfill
  \begin{minipage}{0.221\textwidth}
     \centering 
    \includegraphics[width=\linewidth]{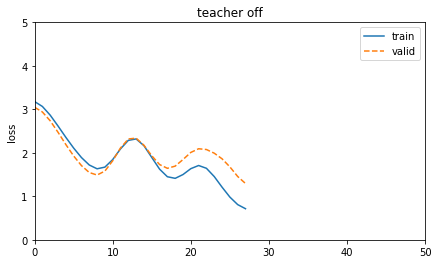}\\
    \includegraphics[width=\linewidth]{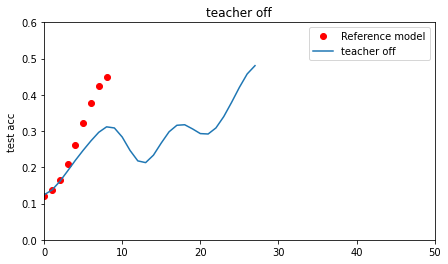}\\
    \includegraphics[width=0.6\linewidth]{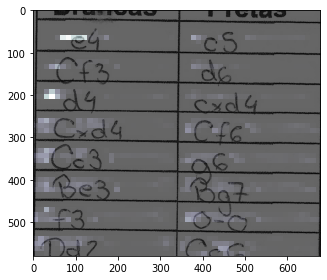}
    
    {\scriptsize (b) Teacher forcing disabled}
  \end{minipage}
  
  \bigskip
  \begin{minipage}{0.221\textwidth}
     \centering 

    \includegraphics[width=\linewidth]{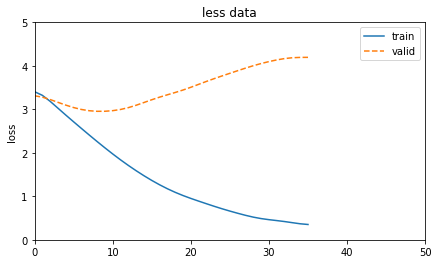}\\
    \includegraphics[width=\linewidth]{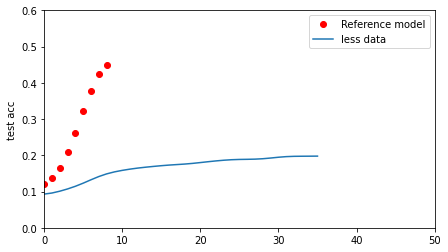}\\
    \includegraphics[width=0.6\linewidth]{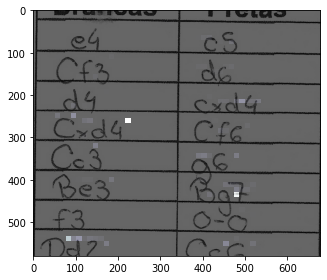}
    
    {\scriptsize (c) Less data (2K)}
  \end{minipage} \hfill
  \begin{minipage}{0.221\textwidth}
     \centering 

    \includegraphics[width=\linewidth]{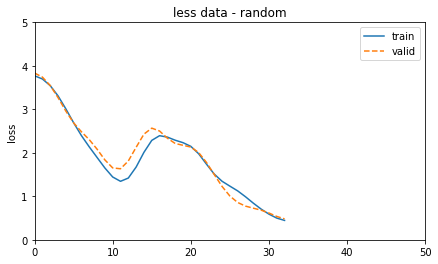}\\
    \includegraphics[width=\linewidth]{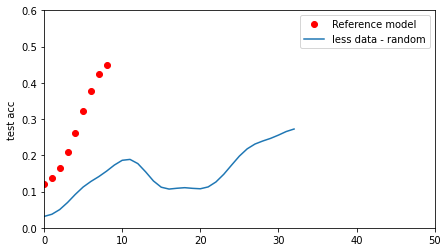}\\
    \includegraphics[width=0.6\linewidth]{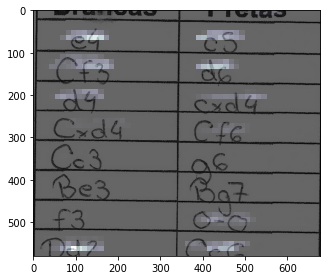}
    
    {\scriptsize (d) Less data (2K), random sequences}
  \end{minipage}

  \bigskip
  \begin{minipage}{0.221\textwidth}
     \centering 
    \includegraphics[width=\linewidth]{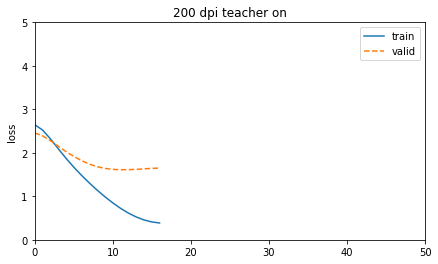}\\
    \includegraphics[width=\linewidth]{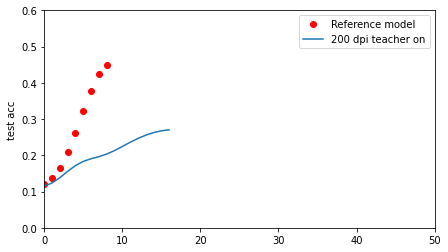}\\
    \includegraphics[width=0.6\linewidth]{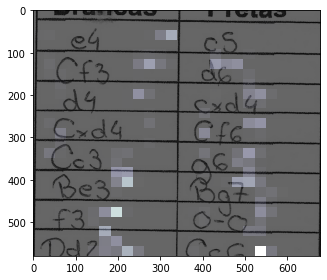}\\
    
    {\scriptsize (e) Smaller image (200dpi)}
  \end{minipage} \hfill
  \begin{minipage}{0.221\textwidth}
     \centering 

    \includegraphics[width=\linewidth]{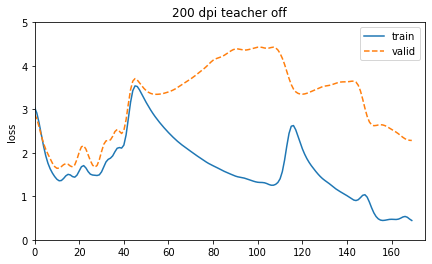}\\
    \includegraphics[width=\linewidth]{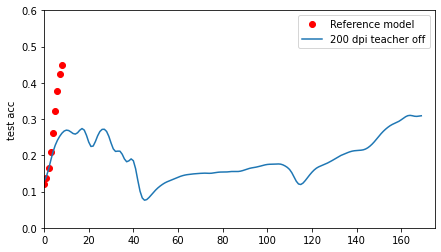}\\
    \includegraphics[width=0.6\linewidth]{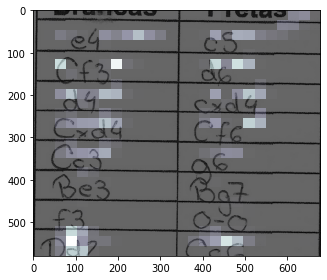}

    {\scriptsize (f) 200dpi, teacher forcing disabled}
  \end{minipage}
    \caption{Training and validation loss curves (top left panel), accuracy curves (top right panel), and attention maps (bottom panel). See text for details.}
    \label{fig:experiments}
\end{figure}

\subsection{Effect of factors in the training of subtasks}

The factors we change in our reference setup (the one in Section~\ref{sec:setup}) are teacher forcing, number of training instances, sequence randomization, and image size. Results are gathered in Fig.~\ref{fig:experiments}. 

Results of the reference setup are shown in Figure~\ref{fig:experiments}(a). Training convergence (top graph) occurs  in less than 10 epochs and without overfitting. Test accuracy increases accordingly (bottom graph). The map shows that attention has been learned. The remaining cases correspond to changing one factor in the reference setup.

\subsubsection*{Teacher forcing}
Figure~\ref{fig:experiments}(b) shows results with teacher forcing turned off. Without teacher forcing, the training convergence is much slower (it takes almost 30 epochs) and there is oscillation in loss and accuracy. This makes sense as teacher forcing allows proper training at later positions (due to the use of ground truth information regarding previous positions), leading thus to efficient training over the entire sequence. In contrast, without teacher forcing the training at later positions suffer with errors in previous steps, being properly trained only when training of previous positions are stabilized.

\subsubsection*{Training set size}
Figure~\ref{fig:experiments}(c) shows the result using  a reduced training set (from 5K to 2K instances). The training loss decreases slowly but consistently. However, there is large overfitting, evidenced by the gap between the training and validation loss curves. In this case, the network did not learn attention. This hints that training convergence is being achieved based on the predictability of the sequence, without relying on visual information.

\subsubsection*{Sequence randomization}
When the sequences are highly predictable, it might be possible that it hinders proper training of the attention layer. Thus, we repeat the last experiment, with 2K training instances, but 'breaking' the predictability (we use artificially generated images with moves placed in random order). Figure~\ref{fig:experiments}(d) shows slow convergence, but this time it is clear that the attention is better trained. As a consequence, there is even a slight increase in test accuracy compared to the previous case. 

\subsubsection*{Image size}
With proper alignment, it is reasonable to assume that the quality of the image features should play a fundamental role in recognition performance. Thus we test a setup using images of half of the resolution of the reference setup. A consequence of this reduction is a proportional reduction in the last feature map size, implying therefore half of attention points per line of the scoresheet. This coarse granularity may result in features that do not capture fine details. As shown in Figure~\ref{fig:experiments}(e), with this setup the resulting model overfits, has smaller accuracy, and its attention map shows that the alignment was reasonably trained, but not as well as in the case of the reference model.

While it is clear that recognition becomes difficult in lower resolution images, based on our understanding of the human visual system, it still should be possible to learn attention. In other words, we are able to focus visual attention to a given point in the scene, even when we are not able to clearly see and discern what is in that point.
To verify if this holds, we show in Fig.~\ref{fig:experiments}(f) the result of training with images of lower resolution (as in Fig.~\ref{fig:experiments}(e)) but now with teacher forcing disabled. As seen, convergence occurs only after 160 epochs of training, which is in sharp contrast compared to less than 20 epochs needed in case (e) or compared to less than 30 epochs in case (b). However, compared to the previous case, the network learned attention better. We argue that by turning off teacher forcing, we make predictability learning more difficult and force the network to also rely on correct alignment (to help recognition). 

\subsubsection*{Sequence size}
\begin{figure}[htb]
\centering
    \includegraphics[width=.35\textwidth]{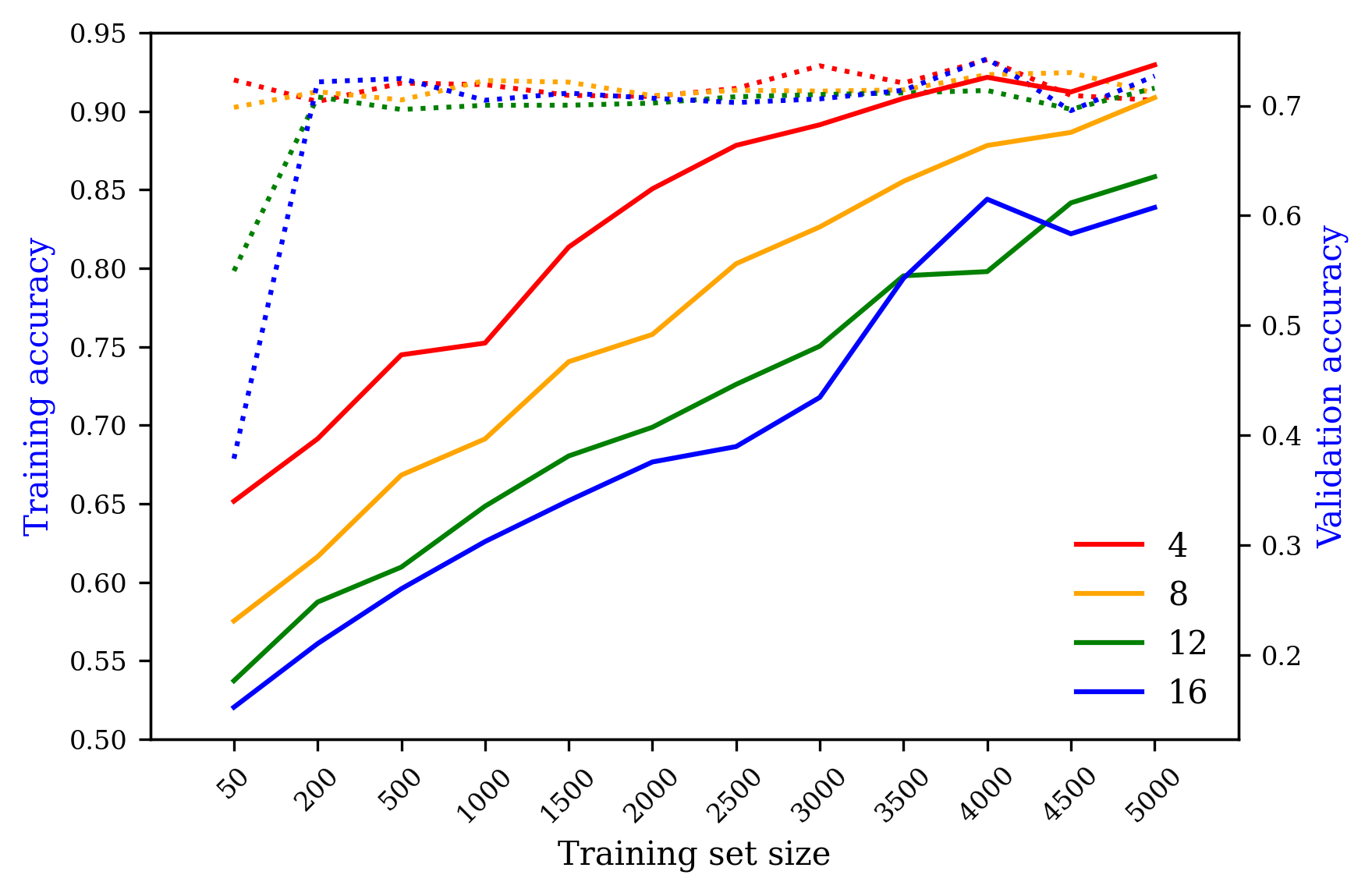}\\
    \includegraphics[width=.35\textwidth]{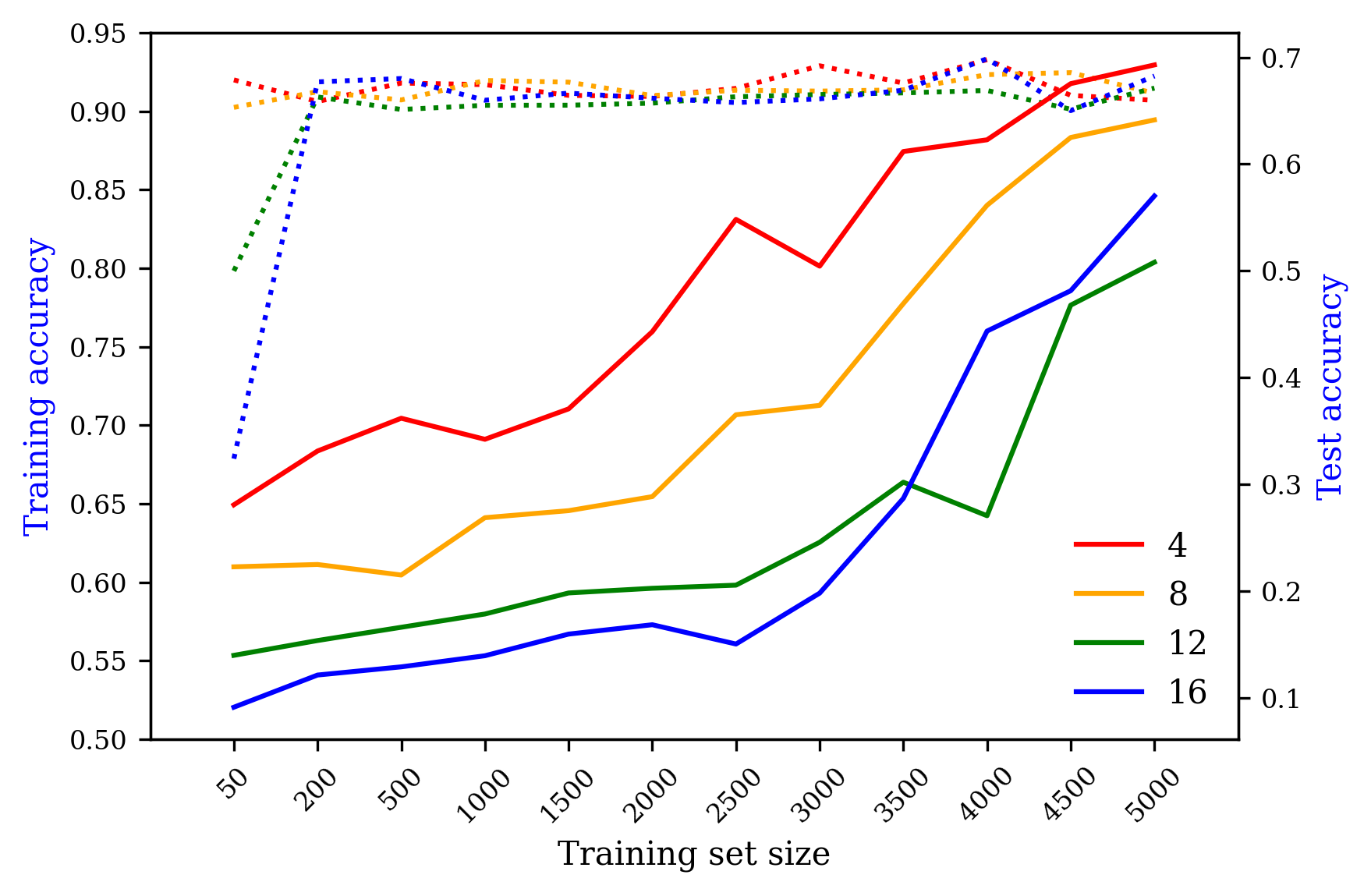}
    \caption{Training, validation and testing accuracy: Varying training set size (indicated in the $x$-axis) and sequence length (indicated by color). Dashed lines refer to training accuracy.}
    \label{fig:length}
\end{figure}

We repeat training using the reference setup for sequences of length 4, 8, 12 and 16, varying the number of training instances. Figure~\ref{fig:length} shows trends of training against validation and training against test accuracy along training set size. Note that accuracies are computed on sequences of different length and thus they are not directly comparable. The first point to be highlighted is the gap between training and validation curves in the top panel graph. While training accuracy easily reaches rate above 90\% (except for very small training sets), validation accuracy only slowly increases as we increase the training set size. Moreover, the larger the sequence, the larger is the gap. Comparing validation and test curves, it is noticeable that test curves start to present clear increase only after some time. In fact, by visual inspection, we observed that attention quality starts to become better around the regions where the test curve exhibits a steep increase. We hypothesize that when that happens the model might be starting to generalize.

\subsection{Discussions}
\label{sec:discussions}

In this section we analyze how learning regarding each of the subtasks are affected by factors and also by other subtasks. 

We start discussing predictability. According to the experiments, when training is performed with sets of highly predictable sequences (meaning few instances or shorter instances), convergence occurs based mostly on predictability only. This fact is supported by some experiments we did in the exploratory step, where we observed that even using a network consisting only of decoder RNN, convergence was achieved when the sequences were ``easily'' predictable. 
In fact, predictability of sequences becomes harder with longer sequences, larger number of instances (diversity of play sequences), or sequences with random moves. In this sense, the amount of data seems to act as a regularizer that prevents the model from specializing only in predictability (language model), forcing the learning of alignment (attention map). Random sequences would be an extreme case of regularization, in which the predictability of the data is totally removed. There is, therefore a competing relation between learning predictability and alignment.

With regard to alignment, the attention layer is a crucial component (this was clearly observed comparing training with and without attention layers during the exploratory phase of this work -- not reported here). Learning the alignment seems to be favored when, in addition to less predictability, the input images have better quality. However, image quality has a much greater effect on the recognition, as we have seen when comparing results of the reference setup and the ones using images with half of the resolution.

Finally, it is clear that without proper alignment, recognition accuracy would be limited by predictability accuracy only. This establishes the dependency relationship between alignment and recognition. However, correct alignment is not sufficient for good recognition. As shown, although alignment can be learned using relatively small resolution images, accuracy gain is obtained only with sufficiently larger resolution images. However, it is important to note that more than the resolution of the input image itself, a determinant aspect is the density of attention points on the input image. This number seems to be crucial for a proper encoding of the visual information for the recognition task. In this sense, the part of the network that is most related to this subtask is the convolutional module. In this work we kept this module frozen. A possible customization of this module is left for future work. The fact that despite using a relatively small amount of writing samples and relatively high granular attention leads the model to achieve reasonable accuracy on test data indicates that recognition is also being helped by predictability. In this sense, we may say that there is a collaboration relation between these two tasks.

\begin{figure}[htb]
\centering

\begin{tikzpicture}[scale=0.9, every node/.style={scale=0.9}]]
      \node (rect) at (2,1) [draw,rounded corners=0.5cm,fill=blue!15,dotted,align=center,minimum width=6.8cm,minimum height=3cm] {\color{blue!50}READING};
      
    \node[draw=blue!15,fill=blue!15,ellipse,name=A,minimum size=1cm] at (0,0) {Alignment};
    \node[draw=blue!15,fill=blue!15,ellipse,name=R,minimum size=1cm] at (4,0) {Recognition};
    \node[draw=blue!15,fill=blue!15,ellipse,name=P,minimum size=1cm] at (2,2) {Predictability};
    
    \node[draw=black,fill=white,ellipse,,minimum size=0.9cm] at (0,0) {Alignment};
    \node[draw=black,fill=white,ellipse,minimum size=0.9cm] at (4,0) {Recognition};
    \node[draw=black,fill=white,ellipse,minimum size=0.9cm] at (2,2) {Predictability};
    \draw[->,line width=0.5mm] (A.east) -- (R.west) node[midway, below] {{\small Dependence}};
    \draw[<->,line width=0.5mm] (P.south east) -- (R.north) node[midway,right] {{\small Collaboration}};;
    \draw[>-<,line width=0.5mm] (A.north) -- (P.south west) node[midway,left] {{\small Competition}};
\end{tikzpicture}
    
    \caption{Subtasks of the reading task and relationship between them.}
    \label{fig:subtasks}
\end{figure}
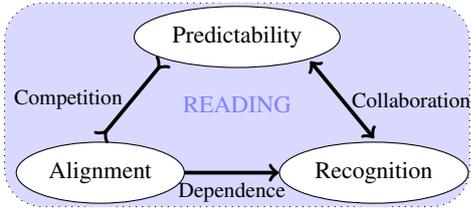

Based on the above analysis, we identify competition, dependence and collaboration relationships among the three subtasks, as summarized in the diagram of Figure~\ref{fig:subtasks}.
Considering a degree of difficulty, the subtasks could be ordered from the easiest to the most difficult, as follows: (1) predictability, (2) alignment, and (3) recognition.

\section{Incremental training}
\label{sec:incremental}

The experiments in the previous section were done with the aim of uncovering the effects of some factors on training quality. As such, the training process was not optimized for final performance. For instance, training was stopped as soon as training loss smaller than 0.25 was achieved. 
The reference model achieved 51.53\% of accuracy on test data, for sequences of length 16. In this section we explore an incremental training strategy. We start with the reference model and further train it, in successive refinement steps. Each step consists in changing the training set and training for some additional epochs.



\begin{figure}[htb]
  \centering
  \includegraphics[width=0.8\linewidth]{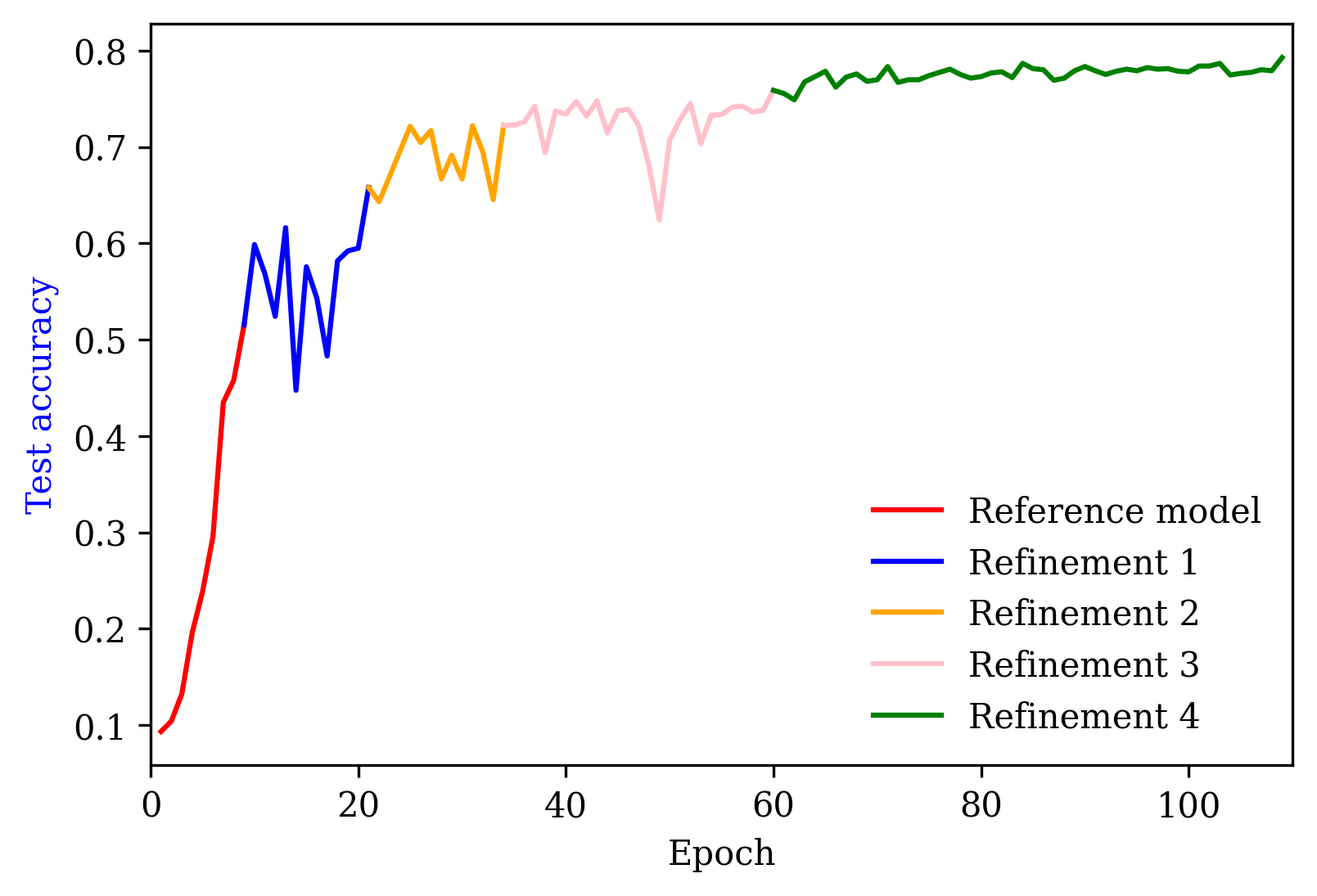}
  \caption{\label{fig:incremental_training}Evolution of test accuracy in an incremental training process. See text for details.}
\end{figure}

Figure~\ref{fig:incremental_training} shows the evolution of the test accuracy along training epochs in the four successive refinement steps. In all of the steps, we included the 2.3K handwritten instances and completed the training set with artificially generated images. Refinement 1 was executed using 10K training instances. After this step, test accuracy increased from 51.53\% to 65.78\%. Considering only the first position, it increased from 70.17\% to 85.08\%. The next two refinement steps were executed using 70K and 10K training instances, and test accuracy increased to 72.25\% and then to 75.87\%, respectively. The last refinement step was executed using only the handwritten instances. At the end, test accuracy reached 79.27\%.

\begin{figure}[htb] 
  \centering
  \includegraphics[width=0.25\textwidth]{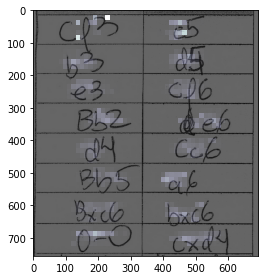}

  \medskip
  {\footnotesize Expected (top) and predicted (bottom) sequences}

  \scalebox{0.89}{\footnotesize \texttt{Nf3 e5 b3 d5 e3 Nf6 Bb2 e6 d4 Nc6 Bb5 a6 Bxc6 bxc6 O-O cxd4}}
    
  \scalebox{0.89}{\footnotesize \texttt{Nf3 \textcolor{red}{\textbf{c}}5 b3 d\textcolor{red}{\textbf{4}} e3 Nf6 B\textcolor{red}{\textbf{e}}2 \textcolor{red}{\textbf{g}}6 d4 Nc6 Bb5 a6 Bxc6 bxc6 O-O cxd4}}


    
      
  \caption{Prediction example for an image from a second tournament, not used in training or testing the final model (for this image, accuracy is 0.75 and CER~\cite{Frinken2014} is 0.1145).}
  \label{fig:predicao--hebraica-1}
\end{figure}

Since scoresheets were collected at one tournament only and then split randomly into training and test sets, it is likely that scoresheets of a same writer are in both sets. Therefore, we also test the final model on a scoresheet of a second tournament. The result is shown in Figure~\ref{fig:predicao--hebraica-1}. The model focus at the right places, as shown in the attention map, and presents good recognition rate, correctly detecting 12 out of 16 moves. Recognition errors are related to confusion between visually similar characters such as ``e'' and ``c'', ``b'' and ``e'', ``g'' and "e". 
This type of error analysis hints to the possibility of employing curriculum learning approaches~\cite{bengio2009curriculum}, in an incremental fashion.

\section{Conclusions}
\label{sec:conclusions}

In this work we explored CNN+RNN (encoder-decoder with attention) network in the context of handwritten chess scoresheet reading, where only few instances of real tournament scoresheet images were available. Our main focus was in understanding what exactly prevents the network from learning. Results are discussed in terms of three implicit subtasks (predictability, alignment and recognition). We show how some factors affect the balanced learning of the three subtasks. Training sets containing highly predictable sequences favors predictability learning, making more difficult to properly train the attention layer. Alignment is, in its turn, crucial for recognition, establishing a dependence relation between the two. Finally, recognition relates both to visual information (image feature quality) and predictability (language context), establishing thus the collaboration relationship between predictability and recognition.  We also explored at what extent simple artificial training data can be used to improve recognition accuracy. Using few real samples and a large amount of synthetic images in a incremental training approach, we managed to achieve 79\% of accuracy on sequences of length 16 on real images. Although the experiments were done with sequences with length up to 16, the observed trends give us indications of what might happen with larger sequences.


Awareness regarding how some factors affect the subtasks could help practitioners to make better informed choices. For instance, instead of blindly augmenting the training set, one could first optimize the image feature extraction component. We believe the analysis presented in this work leads to a better understanding of the inner workings of a network and this type of knowledge could be extended to other complex tasks. As future work, we would like to customize the convolutional layers and devise means to work on full sequences, without limiting sequence length. 

\section*{Acknowledgment}

This work is supported by FAPESP (The S\~{a}o Paulo Research Foundation, Brazil) through grants 2017/25835-9 and 2015/22308-2.

\bigskip

\IEEEtriggeratref{10}



\bibliography{paper_final}

\end{document}